\documentclass[draft]{agujournal2019}
\usepackage{url} 
\usepackage{lineno}
\usepackage[inline]{trackchanges} 
\usepackage{soul}
\usepackage[linesnumbered,ruled,vlined]{algorithm2e}
\usepackage{amsmath}
\draftfalse

\begin{document}

%
%

\title{Enhanced predictive skills in physically-consistent way: Physics Informed Machine Learning for Hydrological Processes }

%
%




\authors{Pravin Bhasme \affil{1}, Jenil Vagadiya \affil{2}, Udit Bhatia \affil{1}}


\affiliation{1}{Civil Engineering Discipline, Indian Institute of Technology, Gandhinagar, India}
\affiliation{2}{Computer Science and Engineering Discipline, Indian Institute of Technology, Gandhinagar, India}




\correspondingauthor{Udit Bhatia}{bhatia.u@iitgn.ac.in}




\begin{keypoints}
\item Model to combine  predictive ability of machine learning with process understanding of physics-based models for hydrological processes.
\item Variants of architecture adapted for uncertainty quantification in both target (streamflow) and intermediate variables (evapotranspiration).
\item Performance gains in prediction of target and intermediate variables; annual water balance analysis reveals physical consistency of model.

\end{keypoints}

%
%

%
%


\begin{abstract}
Current modeling approaches for hydrological modeling often rely on either physics-based or data-science methods, including Machine Learning (ML) algorithms. While physics-based models tend to rigid  structure resulting in unrealistic parameter values in certain instances, ML algorithms establish the input-output relationship while ignoring the constraints imposed by well-known physical processes. While there is a notion that the physics model enables better process understanding and ML algorithms exhibit better predictive skills, scientific knowledge that does not add to predictive ability may be deceptive. Hence, there is a need for hybrid modelling approach to couple ML algorithms and physics-based model in synergistic manner. Here we develop a Physics Informed Machine Learning (PIML) model that combines the process understanding of conceptual hydrological model with predictive abilities of state-of-the-art ML models. We apply the proposed model to predict the monthly time series of the target (streamflow) and intermediate variables (actual evapotranspiration) in the Narmada river basin in India. Our results show the capability of the PIML model to outperform a purely conceptual model ($abcd$ model) and ML algorithms while ensuring the physical consistency in outputs validated through water balance analysis. The systematic approach for combining conceptual model structure with ML algorithms could be used to improve the predictive accuracy of crucial hydrological processes important for flood risk assessment.
\end{abstract}


%
%

\section{Introduction}

Streamflow prediction, flood protection, and water resource management challenges continue to be highly relevant for societal as well as economic security and well-being \cite{butts2004evaluation,niu2021evaluating}.  With the ever-increasing availability of computational resources, diverse spatiotemporal datasets from remotely-sensed and in-situ measurements, and advances in Machine Learning (ML) algorithms, data-driven approaches are being used widely to predict the response of catchments to meteorological forcings, including precipitation and surface temperature \cite{tongal2018simulation,parisouj2020employing,feng2020enhancing}.  Despite recent advances in their predictive performance and ability to handle highly complex spatiotemporal datasets, the success of ML algorithms in the domain applications remains elusive due to limited interpretability \cite{gilpin2018explaining}, physical inconsistencies and persistent issues of equifinality (different model structures and parameterization schemes giving the same end state with equal accuracy) \cite{schmidt2020challenges,beven2006manifesto}. To address interpretability challenges, physics-based models, which encapsulate our domain knowledge and physics principles, including conservation of mass, energy, and momentum through mathematical equations, remain the tool of choice for researchers and practitioners \cite{vieux2004evaluation,beven1989changing}. Physics-based models are broadly categorized as conceptual models and physically-based models. These models involve mathematical and semi-empirical equations with the physical basis \cite{devia2015review}. In addition to meteorological and hydrological observations, these models' predictive performance critically relies on identifying the right set of parameters, which is typically achieved through the process of calibration \cite{fenicia2007comparison}. While there is a notion that physics-based models enable better process understanding, and ML methods lead to better predictive skills, the process understanding that does not add to predictive ability may be specious, and improved predictive skills devoid of physical realism may not generalize to unexpected, yet possible scenarios \cite{ganguly2014toward}. With the growing realization that neither purely ML algorithms nor physics-based models suffice to address the domain-specific challenges, physics guided machine learning approaches are gaining significant attention in scientific and engineering domains, aiming to integrate the data and scientific knowledge in a complementary manner \cite{muralidhar2019physics, wang2017physics, zhang2020physics}. 

 Multiple frameworks, both conceptual and applied, for combining physics with machine learning approaches have been proposed in the literature for diverse applications. For example, \citeA{wagner2016theory} have demonstrated the importance of scientific reasoning for discovering reliable structure, property, and processing models, as well as listed nuances associated with machine learning applications in material science. \citeA{wang2017physics} have proposed a data-driven, physics-informed machine learning (PIML) approach for reconstructing discrepancies in Reynolds-averaged Navier Stokes modeled Reynolds stresses. It is achieved by training existing direct numerical simulation databases through machine learning techniques and applied the PIML approach for different flow conditions such as turbulent flows developed in the square duct for varying Reynolds numbers and flows with massive separation. \citeA{muralidhar2019physics} have demonstrated the application of physics-guided model architecture for predicting drag force acting on each particle in fluid flow. Further, they have incorporated physics-based loss functions to capture physical knowledge, avoid constraint violation, and develop a physics-guided deep neural network model. For seismic response modeling, \citeA{zhang2020physics} have developed a physics-guided convolutional network architecture to predict a building's seismic response. \citeA{faghmous2014theory}, \citeA{reichstein2019deep}, and \citeA{ganguly2014toward} have highlighted the need to have scalable data science methods which can incorporate physical and conceptual understanding for better interpretation for applications in earth and atmospheric sciences. The authors themselves noted that while there exist multiple opportunities in the area of incorporating contextual cues from the domain into machine learning models, the domain-specific challenges (that is, preservation of spatiotemporal structures, confidence and uncertainty representation in predictions) should be accounted for to ensure physical consistency in the outputs. \par
 
 \citeA{karpatne2017physics} and \citeA{jia2019physics} developed physics-guided variants of ML models that guide neural network architectures to simulate lake temperature profiles in a physically consistent manner. Specifically, \citeA{karpatne2017physics} demonstrated how to combine the standard neural network models with the energy conservation laws to outperform purely knowledge-based or data-driven models. These goals were achieved in two ways. First, the outputs from physics-based models were fed into ML architecture. Secondly, they introduced physical knowledge-based constraints in the temporal ML model's objective function (Long Short Term Memory (LSTM) in this case). 
 \citeA{liang2019physics} generated a database through simulations of the physics-based model and then processed it with different data-driven approaches to predict the quantity and quality of surface water in agricultural fields. Recently, \citeA{lu2021streamflow} introduced a physics-informed LSTM model to improve the prediction when data is out of distribution. The physics-based Precipitation Runoff Modeling System (PRMS) model-simulated streamflow and meteorological features were given as inputs to the LSTM for the streamflow prediction. We argue that while these approaches could help to improve the predictive abilities of purely physics-driven or ML models, the gains in terms of interpretability and capability of these hybrid models to capture the intermediate variables realistically are yet to be explored. \citeA{khandelwal2020physics} has proposed the LSTM based architecture for streamflow prediction, which emulates the  Soil and Water Assessment Tool (SWAT), a physically-based semi-distributed model. They used the same inputs as the SWAT model into three LSTMs in the first layer to simulate soil water, evapotranspiration, and snowpack. The outputs obtained from the first layer were combined with the original SWAT inputs to feed into another LSTM to model the streamflow. We note that this is one of a few physics-guided data sciences approaches for hydrological modeling that accounts for intermediate variables (i.e., soil water, evapotranspiration, and snowpack). However, a typical SWAT architecture involves multiple intermediate fluxes (e.g., evapotranspiration, surface runoff, groundwater flow, percolation, lateral flow, groundwater recharge, deep aquifer recharge) and state variables (e.g., soil water, snowpack, deep aquifer storage, shallow aquifer storage) \cite{neitsch2004soil}, which were altogether ignored in their architecture, hence limiting the interpretability of outputs.

\par

In this study, we design a modeling framework to combine the conceptual hydrological model with state-of-the-art ML models to leverage ML algorithms' predictive ability with process understanding of physics-based models in a synergistic manner.  Specifically, we
use the structure of the \textit{abcd} model to identify the input (precipitation, potential evapotranspiration, groundwater storage, and soil moisture), intermediate (actual evapotranspiration), and target variables (streamflow at particular gauge location) at various steps (Figure \ref{fig:1}c). Then, we replace empirical equations of our conceptual model (the $abcd$ model) with  ML algorithms at various steps to identify the relationships between input and output  (both intermediate and target) variables. Our model design is inspired by the fact that conceptual hydrological models involve empirical and semi-empirical equations, which are generally developed and validated for specific basins. Therefore,  their application outside the basin boundaries calls for abundant caution. We argue that the use of ML algorithms to identify the complex relationships between input and output variables at various stages of the modeling process adds flexibility to the models having rigid mathematical structures. Hence, PIML models can generalize to the basins where empirical equations of conceptual models may not generalize. We demonstrate the PIML models' ability to capture the intermediate variables with greater accuracy, which in turn transpire into better model performance to model target variable (streamflow in the present case) compared to conceptual and pure data-driven architectures. Finally, we demonstrate the ability of variants of the PIML models to capture the water balance effectively. We demonstrate the proposed model's applicability on the Narmada River Basin with gauge station at Sandia (Figure \ref{fig:1}d). We use a suite of Machine Learning and Deep Learning architectures commonly used in numerous hydrological and earth sciences applications, including LSTMs \cite{kratzert2018rainfall}, Least absolute shrinkage and selection operator (LASSO) and Ridge Regression \cite{yu2007forecasting,lange2020machine}, Support Vector Regression \cite{deka2014support}, Gaussian Process Regression \cite{sun2014monthly}, and Bayesian LSTMs \cite{lu2021streamflow} for purely ML-based as well as hybrid PIML models. 

We organize the rest of the manuscript as follows: in section 2, we present the brief overview of the conceptual $abcd$ model and ML algorithms used in this study. Details of the proposed PIML model and various evaluation metrics deployed in this study are also discussed. In section 3, we discuss the details of the study area and datasets used in this research. In section 4, we compare physics-based, ML, and PIDS algorithms using metrics discussed in section 2. Finally, we discuss the physical consistency of PIML models by performing a water-balance analysis.

\par

\section{Methods}

This section presents the review of the conceptual \textit{abcd} model and various ML models with underlying equations used in this study. Further, we discuss the architecture of the PIML model. Finally, we present a brief overview of various performance metrics used throughout this study.

\subsection{Review of the $abcd$ model}

The $abcd$ model is a simple Water Balance (WB) conceptual hydrological model proposed by \citeA{thomas1981improved}. While it was originally developed to examine the catchment scale water balances at annual scales, variants of the model have been applied for regional and local hydrological investigations at monthly scales \cite{alley1984treatment, vandewiele1992methodology}. The conceptual model involves the parsimonious yet adequate description of various hydrological processes at catchment scale with the lesser computational cost making it highly popular in operational and research practice \cite{clark2010ancient}. Specifically, the $abcd$ model structure has been widely used for hypotheses testing, model performance evaluation \cite{martinez2010toward, bai2015comparison}, and hydrological uncertainty reduction experiments \cite{li2012reducing} owing to a highly realistic representation of various hydrological processes despite its simple structure. Figure \ref{fig:1}a shows the conceptual representation of the $abcd$ model. The model consists of two storage compartments: soil moisture stoarage and groundwater storage. Inputs required are monthly precipitation $(P_{t})$ and potential evapotranspiration $(PET_{t})$, while output generated is streamflow $(Q_{t})$. This model is calibrated with four parameters, $a$, $b$, $c$, and $d$. The parameter $a$, $(0 < a \leq1)$ controls the amount of direct runoff when soil is unsaturated, and parameter $b$ reflects the upper bound of the total of actual evapotranspiration $(ET_{t})$ and soil moisture $(SM_{t})$ at a given time step. The groundwater recharge $(GR_{t})$ is controlled by parameter $c$, while parameter $d$, decides the amount of groundwater storage $(GW_{t})$ to be converted to groundwater discharge $(GD_{t})$.  The direct runoff and groundwater discharge together generate the streamflow. The model includes two state variables: $W_{t}$ (available water)  and $Y_{t}$. While $W_{t}$ is the sum of precipitation $(P_{t})$ at a given time step and soil moisture $(SM_{t-1})$ at the previous time step, $Y_{t}$ represents the sum of actual evapotranspiration $(ET_{t})$ and soil moisture $(SM_{t})$ at a given time step. Equations \ref{eq:1}-\ref{eq:2}  show the mass balance for soil moisture and groundwater storage compartments. Equations \ref{eq:4}-\ref{eq:6} are typical examples  of parameterized relationships among various variables (See Supplementary Information (SI) for complete set of equations).

\begin{subequations}
\begin{equation}\label{eq:1}
    {SM_{t} + ET_{t} + DR_{t} + GR_{t} = SM_{t-1} + P_{t}}
\end{equation}
\begin{equation}\label{eq:2}
    {GW_{t} + GD_{t} = GW_{t-1} + GR_{t}}
\end{equation}
\begin{equation}\label{eq:3}
    {W_{t} = SM_{t-1} + P_{t}}
\end{equation}
\begin{equation}\label{eq:4}
    {Y_{t} = SM_{t} + ET_{t} = \frac{W_{t} + b}{2a} - \sqrt{\left(\frac{W_{t} + b}{2a}\right)^2 - \frac{b.W_{t}}{a}}}
\end{equation}
\begin{equation}\label{eq:6}
    {ET_{t} = Y_{t} \times (1 - e^{-PET_{t}/b})}
\end{equation}

\end{subequations}

\subsection{Review of Machine Learning Algorithms}

Recent advances in the field of machine learning have provided many methodological opportunities to meet the evolving needs and challenges of hydrological research. ML models have demonstrated superior performance in learning patterns and generalizations as well as extracting patterns from complex streams of geospatial and hydrological datasets \cite{lange2020machine,reichstein2019deep}. ML algorithms form the core of the proposed PIML model. Hence, a review of various methods used in this study is presented to clarify the subsequent sections. \par
Long Short Term Memory (LSTM) is an artificial neural network architecture that has gained popularity for sequential data problems. In the context of hydrology, LSTMs have been used for rainfall-runoff modeling at hourly \cite{xiang2020rainfall}, daily \cite{fu2020deep, cheng2020long}, monthly \cite{cheng2020long} time steps as well as for the improvement in the predictions of the physics-based models \cite{yang2019evaluation}. LSTMs are a special kind of recurrent neural networks (RNNs) capable of learning long-term temporal dependencies. The simple RNNs have an issue of vanishing gradient, which can be removed by LSTMs with the introduction of gates and memory cells \cite{hochreiter1997long}. \par

 Bayesian Neural Networks (BNNs) have advantages over neural networks, such as they can handle small data well while generating uncertainty bounds of predictions. Since BNNs incorporate posterior inference in standard neural networks,  these architectures have gained popularity for uncertainty quantification \cite{marshall2004comparative, yang2007bayesian, raje2012bayesian}. Recently \citeA{lu2021streamflow} has applied Bayesian LSTM for uncertainty quantification in streamflow prediction. BNNs were made by approximating the intractable Bayesian Inference. BayesByBackprop \cite{pmlr-v37-blundell15} is a backpropagation-compatible algorithm to learn probability distribution of a neural network. The same concept is extended to create Bayesian LSTM. BNNs can provide us uncertainty on weights by sampling them from a distribution parameterized by trainable variables \cite{esposito2020blitzbdl}. \par
 
 Gaussian Process Regression (GPR) is a non-parametric kernel-based probabilistic model, which has gained wider popularity in the domain of ML \cite{williams2006gaussian}. Unlike various ML algorithms that determine the exact values of parameters, GPR infers the probability distribution over all possible values using the prior distribution and updated distribution (known as a posterior distribution) that incorporates information from both prior distribution and available data. Researchers have used the GPR for daily \cite{rasouli2012daily} and monthly \cite{sun2014monthly} streamflow forecasting. \par

Support Vector Machine (SVM) is developed for classification, and it is extended to regression by \citeA{vapnik1995nature}. Support Vector Regression (SVR) is one the most popular ML approaches used for daily \cite{dibike2001model, malik2020support} and monthly \cite{maity2010potential} streamflow predictions. SVRs can efficiently learn the non-linear relationships between predictors (input variables) and predictands (output variables) using the kernel trick, which maps the inputs into linearly solvable high-dimensional feature spaces.

\par
The least absolute shrinkage and selection operator (LASSO) \cite{tibshirani1996regression}, and Ridge Regression \cite{hoerl1970Ridge} are some of the simplest techniques which are widely used to reduce model complexity and prevent overfitting. LASSO regression works by reducing the model complexity and feature selection by penalizing the absolute sum of coefficients. As a result of this regularization, some of the coefficients that do not affect the output are reduced to zero. In Ridge regression, the square of the coefficients' magnitude is penalized instead of the absolute sum. When coefficients take large values, the objective function (typically Mean Square Error function) is penalized, resulting in the coefficients' shrinking during the optimization process. LASSO and Ridge regressions have been widely applied for streamflow forecasting, and their outputs have been found comparable with state-of-the-art ML algorithms  \cite{lima2010climate,chokmani2008comparison,xiang2020rainfall}. 
\par

 The details of all the ML models used in this study are presented in SI. 

\subsection{Physics Informed Machine Learning model}

\begin{figure}[!ht]
    \centering
    \includegraphics[width=1.0\textwidth,keepaspectratio]{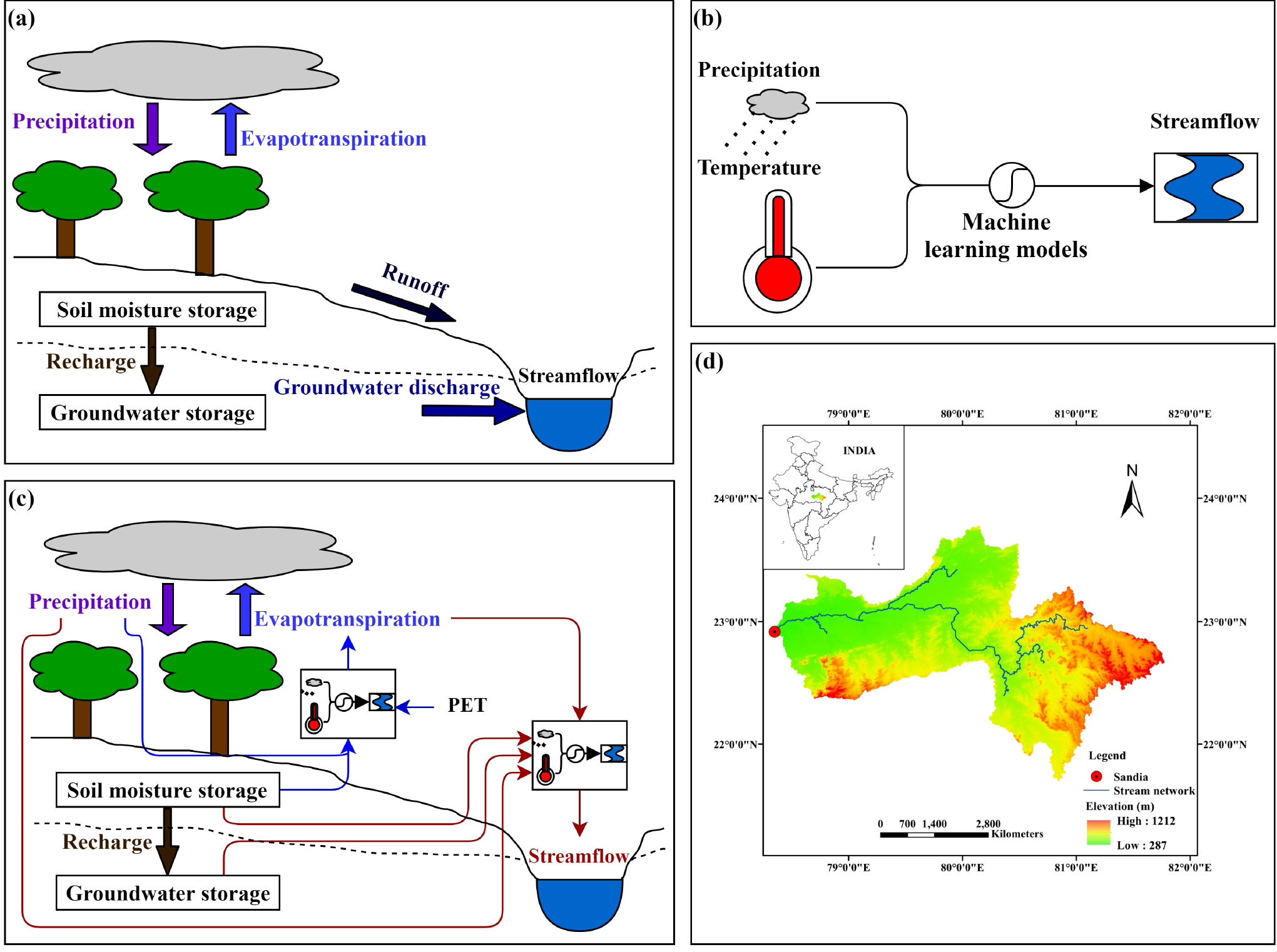} 
    \caption{Illustration of different model architectures used in this study. (a) Conceptual representation of the $abcd$ model. Thick arrows represent the different processes (and directions) of various fluxes in the conceptual model, whereas the rectangular boxes indicate two storage compartments (Soil moisture and groundwater) inside the model. (b) Pictorial representation of ML architecture with precipitation and temperature as input variables (predictors) and streamflow as target variable (predictand). Both linear and non-linear ML algorithms are used here. (c) Proposed PIML model architecture:  the ML approach is embedded into the $abcd$ model architecture. The blue arrows show functional dependencies of actual evapotranspiration (ET) on precipitation (P), soil moisture storage (SM), and potential evapotranspiration (PET). Similarly, red arrows show that evapotranspiration, groundwater, and soil moisture storages would be required predictors for streamflow estimation. (d) The details of the study area. It shows the Sandia gauge station location and digital elevation model of the part of the Narmada river basin considered in this study.}
  \label{fig:1}
\end{figure}

The proposed PIML model provides a way to combine the physics-based conceptual model with various ML approaches to enable better process understanding and improve predictive performance while being mindful of physical consistencies (e.g., water balance). 
The proposed model's premise is as follows: use the covariate structure of the physics-based conceptual model (the $abcd$ model in this case) and replace rigid mathematical relationships among input and output variables at various steps using ML algorithms. While the conceptual model structure provides the physics-informed choice of covariates and interpretable structure, ML algorithms help in extracting complex relationships between the input and output variables. 
For example, in the $abcd$ model, $ET_t$ is non-linear function of $SM_{t-1}$, $PET_t$, and $P_t$ (Equations \ref{eq:3}, \ref{eq:4} and \ref{eq:6}). In the PIML model, we use $SM_{t-1}$, $PET_t$, and $P_t$ as inputs (or predictors) into our embedded ML model and obtain the estimate of $ET_t$ using various ML algorithms (deployed independently) described in Figure \ref{fig:1}c. The estimates of $ET_t$ thus obtained is combined with $SM_t$, $SM_{t-1}$, $GW_t$, $GW_{t-1}$ and $P_t$ to obtain new covariate matrix, which is then fed into next layer of ML algorithm to obtain the final estimates of $Q_{t}$ (target variable in this case). We note that choice of these covariates is governed by the water-balance equation (Equation \ref{eq:43}): 

In general form, the functional relationship for the $ET_{t}$ and $Q_{t}$ can be written as follows.
\begin{subequations}
\begin{equation}\label{eq:43}
    {Q_{t} = f(P_{t},ET_{t},SM_{t},SM_{t-1},GW_{t},GW_{t-1})}
\end{equation} 
\begin{equation}\label{eq:44}
    {ET_{t} = g(P_{t},SM_{t-1},PET_{t})}
\end{equation}
\end{subequations}
The exact function form of $f$ and $g$ is determined by embedded ML models (Figure \ref{fig:1}c). 

\subsection{Evaluation metrics}
For model performance evaluation, we have used Nash-Sutcliffe Efficiency (NSE), Percent Bias (PBIAS), Root Mean Square Error (RMSE). Widely used in various hydrological applications \cite{ najafi2016ensemble, swain2017streamflow, paul2019diagnosing, wagena2020comparison}, these metrics assess model efficiency, biases in the model predictions, and estimate errors in the model outputs, respectively.

\subsubsection{Nash-Sutcliffe Efficiency}
The NSE \cite{nash1970river} is a reliable and widely used statistic to assess goodness of fit for hydrological models \cite{mccuen2006evaluation}. The NSE value has a range of $-\infty$ to 1.0. When the NSE value is 1, it shows a perfect match between modeled output and observed data, while if it is less than 0, it shows observed mean is a better predictor than the model output. Following equation shows the formula for NSE calculation:
\begin{subequations}
\begin{equation}\label{eq:45}
    {NSE = 1 - \frac{\sum_{i=1}^{n}(O_i - S_i)^2}{\sum_{i=1}^{n}(O_i - \bar{O})^2}}
\end{equation}
where, $S_i$, $O_i$, and $\bar{O}$ are model output, observed data, and mean of observed data, respectively. 
\subsubsection{Percent Bias}
It helps to determine how well the model can estimate the average magnitudes of the required output. The PBIAS ranges from $-\infty$ to $\infty$. Its optimal value is 0, while the positive and negative values show model underpredicts and overpredicts, respectively. 
\begin{equation}\label{eq:47}
     PBIAS = \frac{\sum_{i=1}^{n}(O_i - S_i)}{\sum_{i=1}^{n}O_i} \times 100
\end{equation}
\subsubsection{Root Mean Square Error}
It is the measure of deviations in model output from observed data. The RMSE ranges from 0 to $\infty$, and when it is equal to 0, it shows both modeled output and observed data are perfectly match each other.
\begin{equation}\label{eq:46}
    RMSE = \sqrt{\frac{\sum_{i=1}^{n}(O_i - S_i)^2}{n}}
\end{equation}

\end{subequations}

The model performance is evaluated using the criteria outlined in \citeA{moriasi2015hydrologic}. For monthly simulation, $NSE > 0.8$, $0.7 < NSE \leq 0.8$, $0.5 < NSE \leq 0.7$ and, $NSE \leq 0.5$, the model performance is considered as very good, good, satisfactory and unsatisfactory, respectively. Negative NSE values indicates the model performance is unacceptable.

\section{Study area and datasets}

To illustrate the proposed model's applicability, we have selected the part of the Narmada river basin up to gauge station located at Sandia (Figure \ref{fig:1}d). Flowing through the states of Gujarat and Madhya Pradesh, the Narmada River is a 1312 km long river draining a 98796 $km^2$ area. The study area considered here has a drainage area of 38,571 $km^2$.  The daily precipitation and (minimum and maximum) temperature at $0.25^o\times0.25^o$ and $1^o\times1^o$ spatial resolution, respectively,  are obtained from the India Meteorological Department (IMD) for the period of 1979 – 2014. The observed streamflow data of the Sandia gauge station (22.92°N, 78.35°E) is obtained from the India Water Resources Information System (India-WRIS; \textit{https://indiawris.gov.in/}). The daily soil moisture, groundwater storage, and evapotranspiration are obtained from Global Land Data Assimilation System (GLDAS) Catchment Land Surface Model L4 daily datasets, available through the archives of the Goddard Earth Sciences Data and Information Services Center (GES DISC) of National Aeronautics and Space Administration (NASA) $(https://disc.gsfc.nasa.gov/)$. 

\begin{table}
\caption{Datasets used in this study}
\centering
\begin{tabular}{p{6cm} p{2cm} p{5cm}}
  \hline
Data  & Spatial resolution & Source  \\
\hline
Precipitation & 0.25°  & IMD \cite{pai2014development} \\
Minimum and maximum temperature & 1°  & IMD \cite{srivastava2009development} \\
Soil moisture, groundwater storage and actual evapotranspiration  &  0.25° & GLDAS \cite{li2019long} \\
Streamflow  &   Gauge at Sandia   & India-WRIS \\
  \hline
\label{table:1}  
\end{tabular}
\end{table}

We calculate the potential evapotranspiration using the Hargreaves method (equation \ref{eq:48}) \cite{hargreaves1985reference}. The daily \textit{PET} values are summed up to monthly values, followed by spatial averaging over the study region.
\begin{equation}\label{eq:48}
    {PET = 0.0023 \times Ra \times \sqrt{(T_{max} - T_{min})}  \times (T_{avg} + 17.8)}
\end{equation}
where, \textit{PET} is potential evapotranspiration in mm/day     and $Ra$ is the extra-terrestrial radiation in $MJm^{-2}day^{-1}$. $T_{max}$, $T_{min}$ and, $T_{avg}$ are maximum, minimum and average surface temperature in degree Celsius respectively. 

To assess the algorithm's generalizability, we consider distinct training and testing periods while ensuring that the test set does not contain examples from the training sets. Specifically, we use the window of  1979-2008 for training and 2009-2014 for validating the models on testing data. We kept similar training and testing data for the $abcd$ model, ML models, and the PIML model. The additional warm-up period required for the $abcd$ model is selected as 1976-1978. Subsequently, all evaluation metrics (See Results) are calculated on the test set.

\section{Results}
\subsection{Evaluating performance of the physics-based model}
\begin{figure}[!ht]
\centering
\includegraphics[width=1.0\textwidth,keepaspectratio]{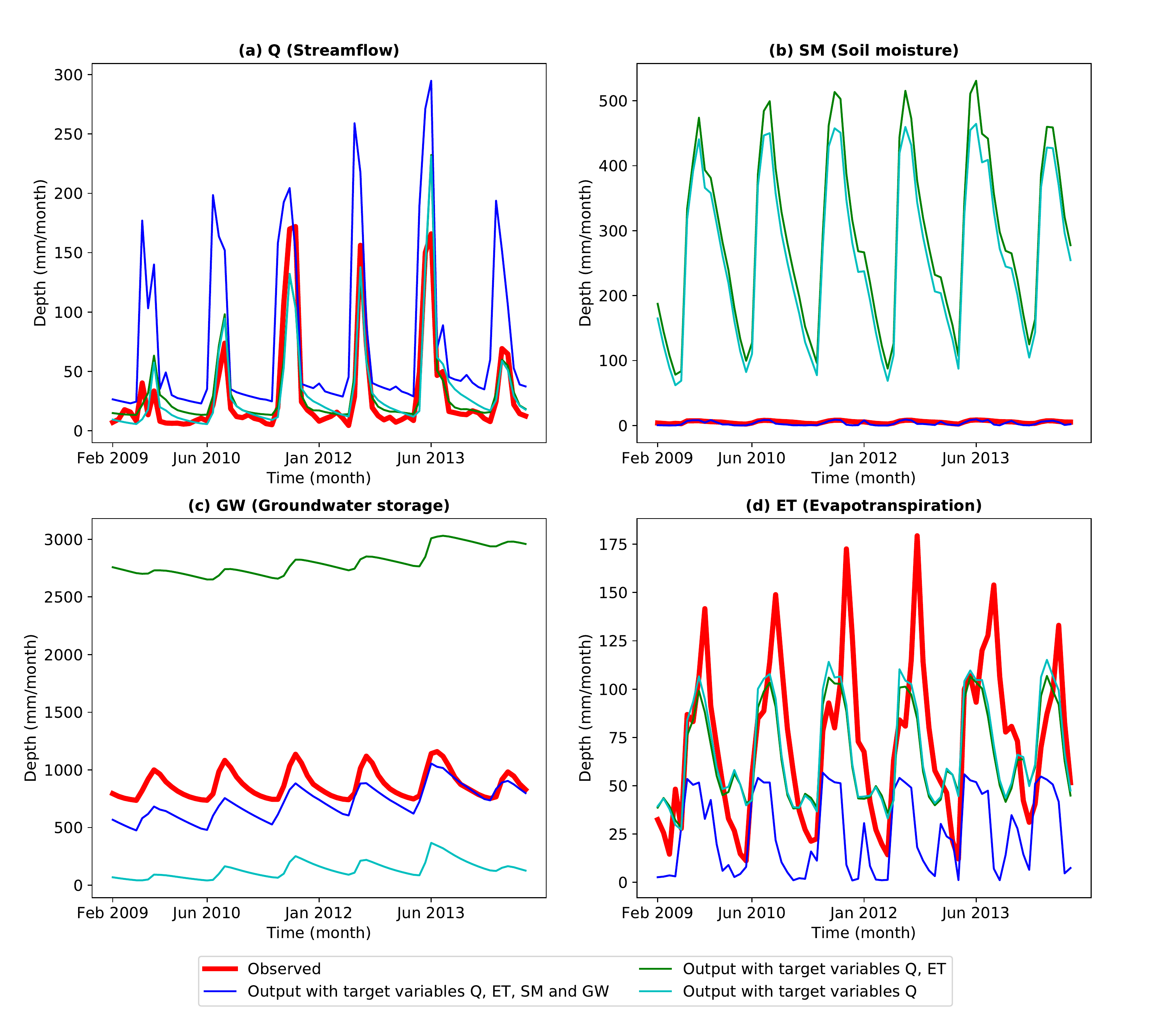}     
\caption{Time-series plots for observed data and model outputs of respective variable: (a) Q (streamflow); Streamflow prediction with model calibrated against Q and ET exhibits a good agreement with observed streamflow. When the model is calibrated against all the four variables (Q, ET, SM, GW), a dip in the performance is noted; (b) SM (soil moisture); When model calibrated against four variables Q, ET, SM, GW, the modeled SM is comparable to observed SM. For other two cases, model cases, calibrated model fails to capture the trend as well as magnitude; (c) GW (groundwater storage); For all cases, significant offset between observed and predicted values of GW time-series is noted  (d) ET (evapotranspiration); same as (a) but for Actual ET.}

\label{fig:2}
\end{figure}

Here, we evaluate the performance of the $abcd$ model in predicting target and intermediate variables. We have considered the three cases. In each case, we use the same model structure but calibrate: (a) modeled streamflow (Q) against observed Q; (b) modeled Q and ET against observed Q and ET, respectively; and (c) modeled Q, SM, GW, and ET against observed Q, SM, GW, and ET, respectively. For all these cases, we use particle swarm optimization to estimate the parameters.  The model warm-up and calibration period are selected as 1976-1978 and 1979-2008, respectively. We test the calibrated model's performance for all cases using monthly data from 2009-2014 (same as the period of testing data used in all ML algorithms throughout the study). Results for all three cases are shown in Figure \ref{fig:2}. Table \ref{table:2} summarizes the calibrated $abcd$ model's performance in terms of key performance metrics for the three cases. The model performance is evaluated using the criteria outlined in \citeA{moriasi2015hydrologic}. For cases (a) and (b), model performance is categorized as "very good."
In contrast, for case(c), the performance is classified as unacceptable in modeling the streamflow (based on the values of NSE). However, in case (b), despite calibrating the model for both Q and ET, the predictive performance on intermediate variable (ET) remains "unacceptable." Further, for case (c), we observe that the model performance was "unacceptable" for all the predicted variables, including streamflow.  Our results imply that while it is possible to obtain remarkable performances on target variables through the calibration process, even simple conceptual models may not simulate various intermediate processes consistently, which are otherwise important for interpretability  (as observed in case (a)). The results thus obtained underscore the need to seriously consider model structure, generalizability, and intermediate process representations in physics-based models \cite{kirchner2006getting}.

\begin{table}
\caption{Performance assessment during validation period (2009-2014) for ET, Q, SM and GW in the $abcd$ model with model calibration against different variables }
\label{table:2}
\centering
\begin{tabular}{p{4cm} p{1.1cm} p{1.1cm} p{1.8cm} p{1.1cm} p{1.1cm} p{1.3cm}}
\hline
Variable & & ET & & & Q & \\
\hline
 Performance metric & RMSE & PBIAS & NSE & RMSE & PBIAS & NSE \\
\hline
 (a) Target: Q & 30.443 & -6.495 & 0.438 & 17.767 & 8.392 & 0.815 \\
\hline
 (b) Target: Q and ET & 31.519 & -10.381 & 0.397 & 17.172 & 11.296 & 0.827 \\
\hline
 (c) Target: Q, ET, SM and GW & 60.860 & -65.289 & -1.247 & 64.303 & 145.771 & -1.423 \\
\hline
\hline
Variable & & SM & & & GW & \\
\hline
 Performance metric & RMSE & PBIAS & NSE & RMSE & PBIAS & NSE \\
\hline
 (a) Target: Q & 286.336 & 4862.764 & -23967.799 & 739.475 & -84.398 & -37.014 \\
\hline
 (b) Target: Q and ET & 314.685 & 5384.296 & -28948.883 & 1934.970 & 221.640 & -259.286 \\
\hline
 (c) Target: Q, ET, SM and GW & 2.776 & -35.066 & -1.253 & 180.558 & -17.592 & -1.266 \\
\hline
\end{tabular}
\end{table}

\subsection{Performance evaluation of ML Algorithms}

\begin{figure}[!ht]
    \centering
    \includegraphics[width=1.0\textwidth,keepaspectratio]{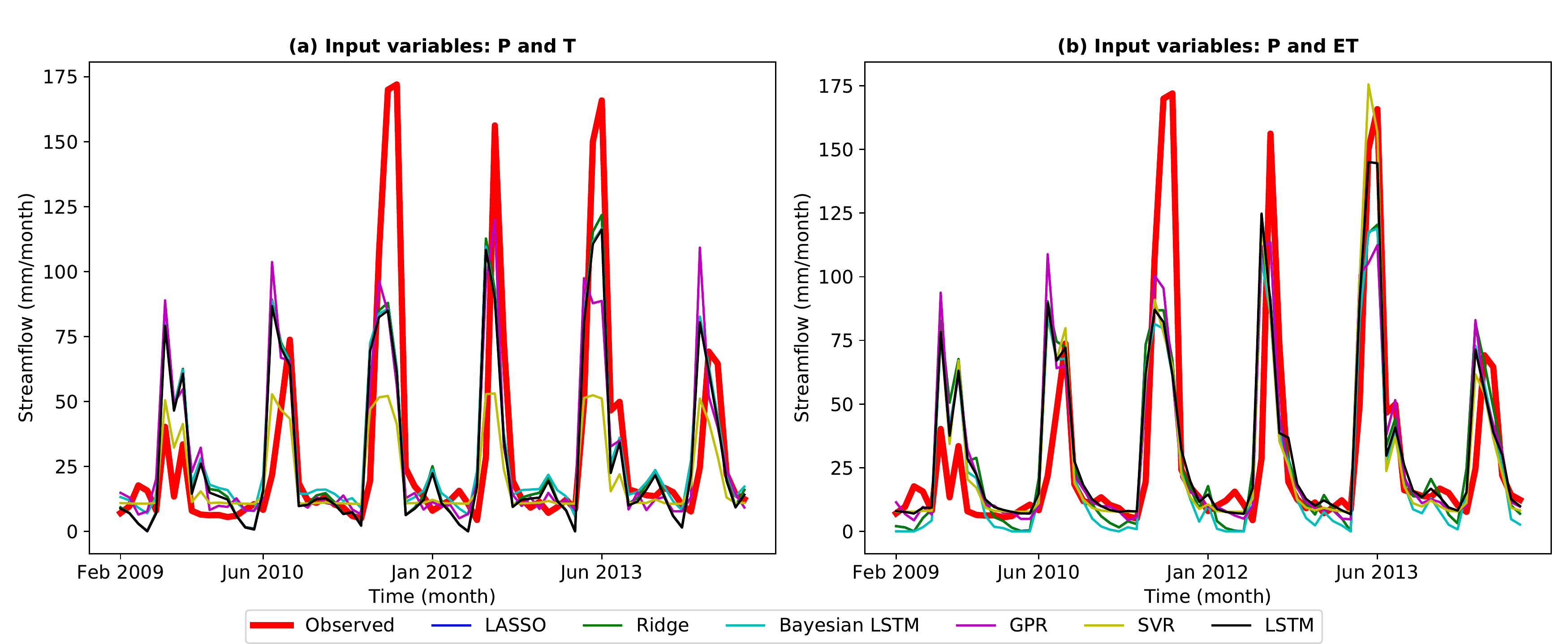}     
  \caption{Time series plots of observed streamflow and modeled output of different ML models with different input variables (a) precipitation (P), average temperature (T) and, (b) precipitation (P), evapotranspiration (ET).}
  \label{fig:3}
\end{figure}

Before we discuss the results of the PIML model, we present the performance of various ML algorithms to model streamflow with a different set of input variables. These ML algorithms would be embedded into the structure of the conceptual model. Thus, it is imperative to understand the ability of ML models to capture relatively straightforward rainfall-runoff relationships. Specifically, we use (a) inputs as precipitation and average temperature and (b) inputs as precipitation and actual evapotranspiration. Figure \ref{fig:3}a and \ref{fig:3}b shows the comparison of all the ML models with observed streamflow for both cases.  Performance metrics for all the algorithms are summarized in Table \ref{table:3}. 

For case (a), LSTM, LASSO, Ridge, Bayesian LSTM exhibit satisfactory performance, whereas performance SVR and GPR lie in the "unsatisfactory" range. Similarly, for case (b), we note that all the models' performance is in the "satisfactory" range based on the NSE criteria. Though Bayesian LSTM shows higher PBIAS,  the improved NSE and decreased RMSE highlight the improvements in the model prediction, which can be attributed to input variables' choice. Further, we adapt the architecture of Bayesian LSTMs to quantify the associated epistemic uncertainty (Fig. 4). We note that despite various ML models' ability to capture the non-linear relationship between predictors and predictands and their scalability, it is often difficult for users to comprehend why specific predictions are made. Thus, there is an opportunity to combine ML and Physics-Based models' architectures in a complementary way to address the associated limitations. In this study, we achieve this by using the $abcd$ model structure and embed ML algorithms in the proposed PIML model to simulate intermediate processes.  
\par
        
\begin{figure}[!ht]
    \centering
    \includegraphics[width=1.0\textwidth,keepaspectratio]{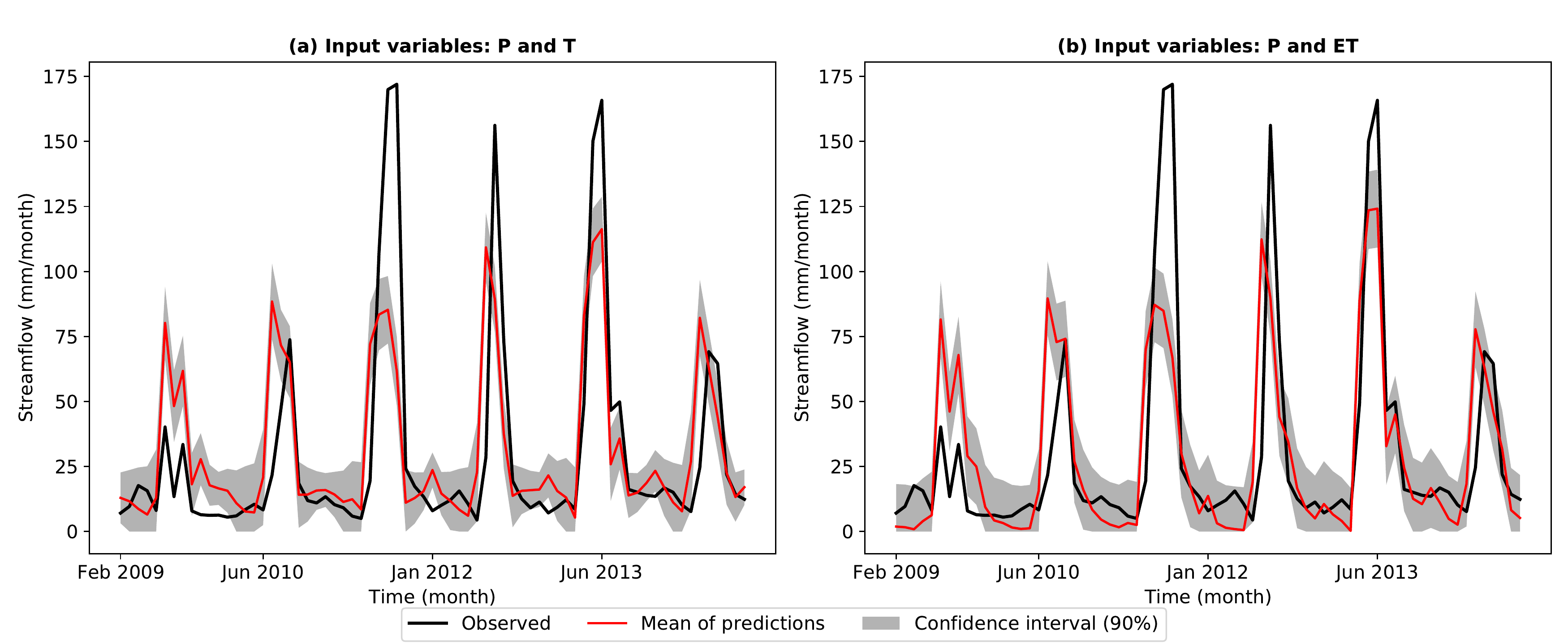}     
  \caption{Time series plots of observed streamflow and modeled output of Bayesian-LSTM model at 90 percent confidence interval with different input variables (a) precipitation (P), average temperature (T) and, (b) precipitation (P), evapotranspiration (ET). The plot shows there is no reduction in uncertainty on comparing case (a) to case (b) even after changing the predictor variables.} 
  \label{fig:4}
\end{figure}
\begin{table}
\caption{Performance assessment of ML models during the testing period (2009-2014)}
\label{table:3}
\centering
\begin{tabular}{p{3.2cm} p{1.1cm} p{2.0cm} p{1.0cm} p{1.1cm} p{2.2cm} p{1.0cm}}
\hline
Input variables & & (a) P and T & & & (b) P and ET  \\
\hline
 Performance metric & RMSE & PBIAS & NSE & RMSE & PBIAS & NSE \\
\hline
LSTM & 28.104 & -6.721 & 0.537 & 27.252 & 4.322 & 0.565 \\
\hline
LASSO & 28.037 & -2.624 & 0.539 & 27.899 & -0.312 & 0.544 \\
\hline
Ridge & 28.041 & -2.638 & 0.539 & 27.902 & -0.307 & 0.544 \\
\hline
SVR & 33.420 & -31.625 & 0.346 & 27.098 & -1.003 & 0.570 \\
\hline
GPR & 30.008 & -2.056 & 0.472 & 27.600 & 1.107 & 0.554 \\
\hline
Bayesian LSTM & 28.198 & 5.684 & 0.534 & 27.730 & -15.230 & 0.549 \\
\hline
\end{tabular}
\end{table}
\subsection{Performance Evaluation of PIML model}

\begin{figure}[!ht]
    \centering
    \includegraphics[width=1.0\textwidth,keepaspectratio]{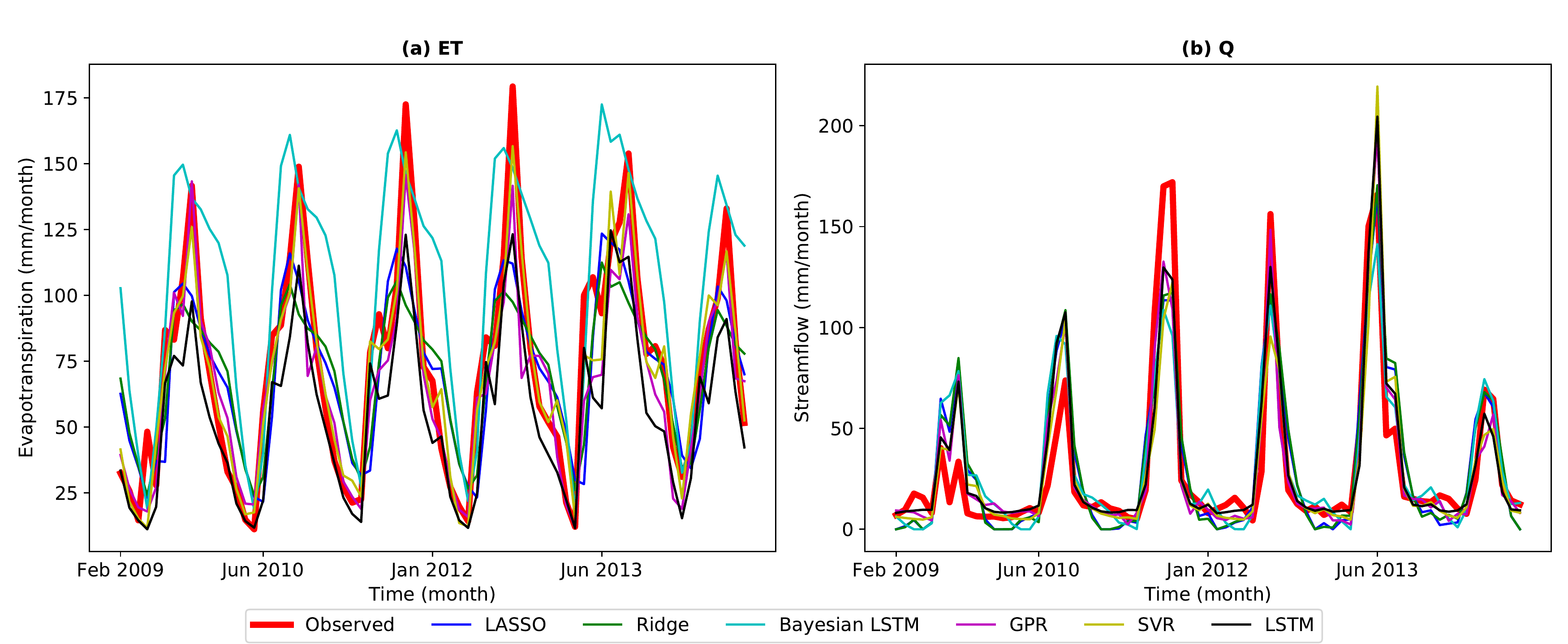}     
  \caption{Time-series plots of PIML model output for (a) ET  and (b) Q.  Most PIML variants are able to capture trends and magnitude with reasonable performance for both the variables. For ET, we note the superior predictive performance of PIML model embedded with SVR. However, notable lag is observed in predictions obtained from PIML+Bayesian LSTM when compared to observations. For Q, all variants are consistent in their predictive performance. PIML with GPR and LSTM outperform other variants. }
  \label{fig:5}
\end{figure}

We evaluate the performance of the PIML model for an intermediate variable (Actual ET) and streamflow (Q) for which credible observations are available for both training and testing periods(Figure \ref{fig:5}a and \ref{fig:5}b). Table \ref{table:4} summarizes the results of the PIML models embedded with various ML algorithms. The NSE value for Q shows that PIML models embedded with LSTM and GPR perform "very good" while other models exhibit "good" performance. We compare the performance of the PIML approach with the corresponding ML approach. For example, the performance of the $abcd$ model embedded with GPR and purely ML-based GPR algorithm are compared. We note consistent improvement in the NSE values obtained from monthly Q for all 6 cases of PIML (Table \ref{table:4}) compared to ML algorithms' performance (Table \ref{table:3}). Further, we note that the best performing PIML architectures ($abcd$+GPR, and $abcd$+LSTM) even outperform the $abcd$ model (Table \ref{table:2}). Further, we calculate the model performance on predicting ET and note that 2 (PIML+SVR), (PIML+GPR) out of 6 variants of PIML exhibit "very good" performance, whereas three variants (PIML+LASSO, PIML+Ridge, PIML+LSTM) can be classified as satisfactory. These five models also outperform the $abcd$ model in predicting monthly ET, thus highlighting the superiority of PIML models in predicting target and intermediate variables. \par

\begin{figure}[!ht]
    \centering
    \includegraphics[width=1.0\textwidth,keepaspectratio]{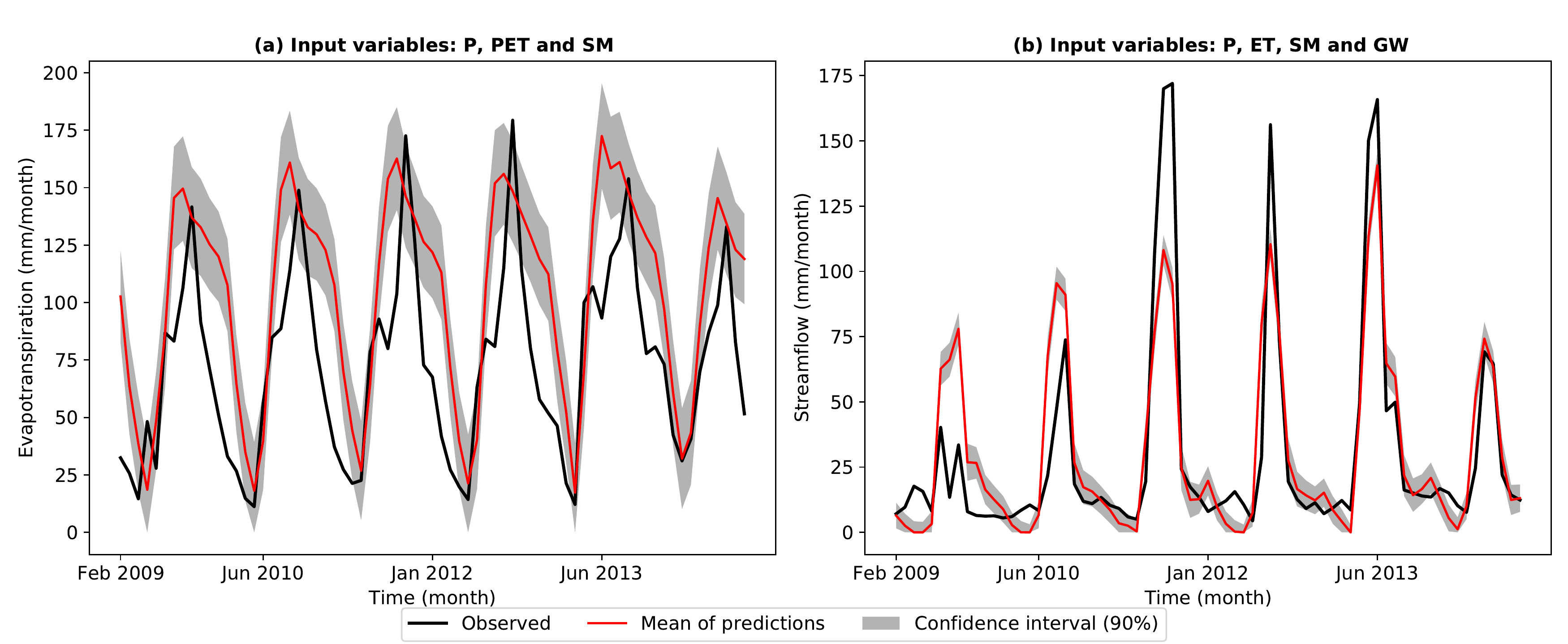}     
  \caption{90 \% confidence interval bounds in the prediction of (a) ET and (b) Q as obtained from PIML model embedded with Bayesian LSTMs. For Q, considerable reduction in uncertainty bounds is reported in comparison to pure ML algorithms} 
  \label{fig:6}
\end{figure}

Further, we demonstrate how PIML approaches can be tailored to quantify uncertainties in the predictions of intermediate and target variables. Specifically,  we quantify the uncertainties in modeled ET and Q using $abcd$+ Bayesian LSTM variant of PIML. Predictions of Q and ET with 90 percent confidence interval are shown in Figure \ref{fig:6}a and Figure \ref{fig:6}b, respectively, for the testing period. In addition to the superior performance of the PIML model, we also notice the reduction in uncertainty bounds in predictions of Q compared to purely ML-based algorithms (Figure \ref{fig:4}b). 

As noted earlier, in addition to improved prediction accuracy over pure physics or ML-based approaches, PIML approaches should generate outputs consistent with physical laws such as conservation of mass. In the context of hydrology, it is imperative to assess the model's ability to simulate annual water balance realistically. 

To check annual water balance, the sum of precipitation should be equal to actual ET, streamflow, and change in soil moisture and groundwater storage. For our study area, soil moisture and groundwater storage changes are observed as -0.045 mm and -0.155 mm, respectively. As these values are negligible in comparison to other terms in the water balance equation, the changes in storage are thus ignored \cite{szilagyi2020water}. 

Here, we consider three variants of PIML: (a) PIML with SVR; (b)PIML with GPR; and (c) PIML with SVR and GPR. While PIML with GPR performs the best in predicting Q, PIML with SVR outperformed other variants in predicting ET. Hence, we experiment with the hybrid variant: SVR for modeling ET in the first layer and GPR to model Q in the second layer in our PIML architecture. It is noteworthy that P, ET, and Q are obtained from three sources (Table \ref{table:1}). Therefore, an initial difference of -9.884 mm is observed between P and the sum of ET and Q. 
We calculate the percentage deviation in ET+Q for all the variants, with observations taken as a benchmark. Variant (c) exhibits  deviation of -1.893 $\%$, whereas variants (a) and (b) have the deviation of -3.255$\%$ , -6.754$\%$ respectively in annual water balance (Table \ref{table:5}). The lower values of $\%$ deviation for three cases demonstrate the ability of PIML approaches to simulate annual water balance consistently.

\begin{table}
\caption{Performance assessment of PIML models during the testing period (2009-2014)}
\label{table:4}
\centering
\begin{tabular}{lllllll}
\hline
Variable & & ET & & & Q  & \\
\hline
 Performance metric & RMSE & PBIAS & NSE & RMSE & PBIAS & NSE \\
\hline
LSTM & 22.704 & -23.966 & 0.687 & 16.083 & 4.124 & 0.848 \\
\hline
LASSO & 25.975 & -3.303 & 0.591 & 20.901 & 3.796 & 0.744 \\
\hline
Ridge & 27.023 & -4.013 & 0.557 & 21.02 & 4.259 & 0.741 \\
\hline
SVR & 11.703 & -2.543 & 0.917 & 19.719 & -4.975 & 0.772 \\
\hline
GPR & 16.407 & -9.287 & 0.837 & 15.001 & -0.626 & 0.868 \\
\hline
Bayesian LSTM & 41.819 & 41.635 & -0.061 & 21.242 & 2.572 & 0.736 \\
\hline
\end{tabular}
\end{table}
\begin{figure}[!ht]
    \centering
    \includegraphics[width=1.0\textwidth,keepaspectratio]{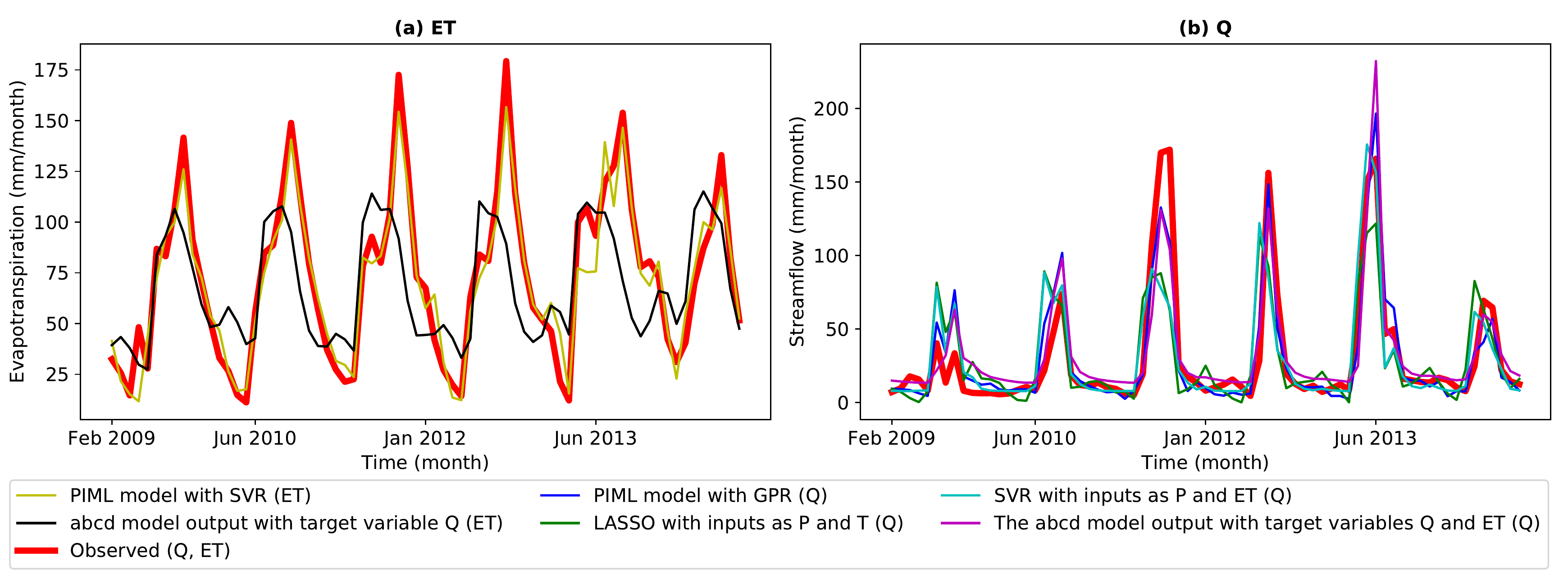}     
  \caption{Time series plots of observed streamflow and evapotranspiration with modeled outputs of the best of all the models (the abcd model, ML models and PIML models):  (a) evapotranspiration (ET) and (b) streamflow (Q).} 
  \label{fig:7}
\end{figure}

\begin{table}
\caption{Annual water balance analysis of physics-informed data science model for testing period (2009-2014). All the values are in mm.}
\label{table:5}
\centering
\begin{tabular}{p{3.0cm} p{1.2cm} p{1.2cm} p{1.2cm} p{1.3cm} p{2cm}}
\hline
 Model/observed & P & ET & Q & ET + Q & Percentage of deviation\\
\hline
Observed & 1231.783 & 878.549 & 363.118 & 1241.667 & 0 \\
\hline
PIML with SVR & 1231.783 & 856.204 & 345.053 & 1201.257 & -3.255 \\
\hline
PIML with GPR & 1231.783 & 796.954 & 360.847 & 1157.801 & -6.754 \\
\hline
PIML with SVR and GPR & 1231.783 & 856.204 & 361.962 & 1218.166 & -1.893\\
\hline
\end{tabular}
\end{table}

\section{Conclusion}

Physics-guided and physics-informed data science approaches have received significant attention in the recent past \cite{muralidhar2019physics, wang2017physics, zhang2020physics, karpatne2017physics}. Several architectures in disparate fields have been proposed to improve the predictive abilities of pure physics-based and ML approaches in a physically consistent manner. Hydrological modeling is a non-linear and complex problem with ample scope for improvement. In this study, we propose the PIML model for predicting target as well as intermediate variables. We demonstrate the applicability of this model on a single hydrological unit to predict monthly time-series of actual evapotranspiration and streamflow, the two key variables in hydrological processes \cite{xiong2019identifying}. 

The proposed PIML model is the first-of-its-kind to provide an intuitive way to combine physically interpretable architectures of lumped hydrological models with state-of-the-art ML algorithms in a meaningful way. We also study the ability of these models to quantify uncertainties in both intermediate and target variables. For our study area, outputs from the PIML model exhibited high prediction performance compared to pure physics-based or ML approaches. Besides, we observe a significant reduction in the uncertainty bounds for predictions of both variables. We also assessed the ability of the PIML models to simulate the annual water budget and noted consistent performance for various variants. 

The PIML model presented in this manuscript allows us to capture complex and non-linear dependencies between hydrological variables while being mindful of the logical sequence which the physics-based model guides. The proposed approach provides a way to add flexibility to otherwise rigid model structures of state-of-the-art hydrological models widely used in hydrology. While we demonstrate the applicability of this approach using a simple WB model ($abcd$ model), this approach can be extended to other conceptual models, which may involve a much larger number of intermediate steps.

Future extensions to the PIML framework for hydrological applications need to be validated for distributed as well as semi-distributed model structures, as well as for daily and sub-daily time-steps. Moreover, we have ignored the role of upstream reservoirs, which might impact the predictive skills of models, especially at daily and sub-daily scales. Also, the performance of the PIML model is highly sensitive to the choice of ML algorithms used at various intermediate steps.  As of now, there are no proven guidelines on the best ML model selection. Despite these outlined limitations, the application of the proposed framework can be extended to various scientific and engineering problems such as early warning systems, risk and reliability assessment for hydraulic structures, and flood management.

\acknowledgments
Funding for the project is provided by Scheme for Transformational and Advances in Sciences of Ministry of Education implemented by Indian Institute of Science, Bangalore (Research Project ID: 367 titled 'Physics Guided Data Science Approach for Predictive Understanding of
Hydrological Processes'). The authors thank Professor Auroop Ganguly from Northeastern University, Boston for helpful discussions, and IIT Gandhinagar colleagues Professor Nipun Batra, and Divya Upadhyay for comments on the manuscript.


%
%

%
%
%
%
%

\end{document}


%
%


\title{Supporting Information for "Enhanced predictive skills in physically-consistent way: Physics Informed Machine Learning for Hydrological Processes"}
%
%

%
%



\authors{Pravin Bhasme \affil{1}, Jenil Vagadiya \affil{2}, Udit Bhatia \affil{1}}


\affiliation{1}{Civil Engineering Discipline, Indian Institute of Technology, Gandhinagar, India}
\affiliation{2}{Computer Science and Engineering Discipline, Indian Institute of Technology, Gandhinagar, India}

%
%

%

\begin{article}

%
%

\noindent\textbf{Contents of this file}
\begin{enumerate}
\item Text S1 to S6
\end{enumerate}



\noindent\textbf{Text S1: Decriptive Equations of the $abcd$ model}

\begin{subequations}
\begin{equation}\label{eq:1}
    {SM_{t} + ET_{t} + DR_{t} + GR_{t} = SM_{t-1} + P_{t}}
\end{equation}
\begin{equation}\label{eq:2}
    {GW_{t} + GD_{t} = GW_{t-1} + GR_{t}}
\end{equation}
\begin{equation}\label{eq:3}
    {W_{t} = SM_{t-1} + P_{t}}
\end{equation}
\begin{equation}\label{eq:4}
    {Y_{t} = SM_{t} + ET_{t} = \frac{W_{t} + b}{2a} - \sqrt{\left(\frac{W_{t} + b}{2a}\right)^2 - \frac{b.W_{t}}{a}}}
\end{equation}
\begin{equation}\label{eq:5}
    {SM_{t} = Y_{t} \times e^{-PET_{t}/b}}
\end{equation}
\begin{equation}\label{eq:6}
    {ET_{t} = Y_{t} \times (1 - e^{-PET_{t}/b})}
\end{equation}
\begin{equation}\label{eq:7}
    {DR_{t} = (1 - c) \times (W_{t} - Y_{t})}
\end{equation}
\begin{equation}\label{eq:8}
    {GR_{t} = c \times (W_{t} - Y_{t})}
\end{equation}
\begin{equation}\label{eq:9}
    {GW_{t} = \frac{(GW_{t-1} + GR_{t})}{1+d}}
\end{equation}
\begin{equation}\label{eq:10}
    {GD_{t} = d \times GW_{t}}
\end{equation}
\begin{equation}\label{eq:11}
    {Q_{t} = DR_{t} + GD_{t}}
\end{equation}
\end{subequations}

\noindent\textbf{Text S2: LSTM}

Long Short Term Memory (LSTM) Networks are being used increasingly for various hydrological applications including rainfall-runoff modeling at hourly \cite{xiang2020rainfall}, daily (\citeA{fu2020deep}, \citeA{cheng2020long}), monthly \cite{cheng2020long} time steps as well as for the improvement in the predictions of physics-based models \cite{yang2019evaluation}. LSTMs are special kind of recurrent neural networks (RNNs) capable of learning long-term temporal dependencies. The simple RNNs have an issue of vanishing gradient, which is removed by LSTMs  with the introduction of gates and memory cells \cite{hochreiter1997long} . These gates control the amount of information provided to the memory cells.  The $w_{i}$ and $b_{i}$  denote weights and biases, respectively. The forget gate decides how much past state memory is retained by applying a $sigmoid$ function to current step input, $x_{t}$ and previous step output, $h_{t-1}$. The input gate decides how much input is sufficient to remember by operating a $sigmoid$ function over $x_{t}$ and $h_{t-1}$. The candidate of the new cell state is calculated by a $tanh$ function with $x_{t}$ and $h_{t-1}$. The forget gate and input gate outputs are multiplied by the previous step cell state $C_{t-1}$ and the candidate of the new cell state $\tilde{C_{t}}$, respectively, and their sum updates the cell state. The output gate operates a $sigmoid$ function with $x_{t}$  and $h_{t-1}$ and determines the required memory to generate output. The final output of the LSTM cell is obtained by the product of the output gate's output and a $tanh$ function of the cell state.The detail description  of LSTMs can be found in \cite{xiang2020rainfall}.

\begin{subequations}
\begin{equation}\label{eq:12}
    {f_{t} = \sigma(W_{f}.[h_{t-1},x_{t}] + b_{f})}
\end{equation}
\begin{equation}\label{eq:13}
    {i_{t} = \sigma(W_{f}.[h_{t-1},x_{t}] + b_{i})}
\end{equation}
\begin{equation}\label{eq:14}
    {\tilde{C_{t}} = tanh(W_{c}.[h_{t-1},x_{t}] + b_{c})}
\end{equation}
\begin{equation}\label{eq:15}
    {C_{t} = f_{t} \times C_{t-1} + i_{t} \times \tilde{C_{t}}}
\end{equation}
\begin{equation}\label{eq:16}
    {o_{t} = \sigma(W_{f}.[h_{t-1},x_{t}] + b_{o})}
\end{equation}
\begin{equation}\label{eq:17}
    {h_{t} = o_{t} \times tanh(C_{t})}
\end{equation}
\end{subequations}
\noindent\textbf{Text S3: Bayesian LSTM using Bayes by backprop}

 Bayesian Neural Networks have advantages over neural networks, such as they can handle small data well and give uncertainty bounds of predictions. There are several attempts made in the past for uncertainty quantification using the Bayesian framework (\citeA{marshall2004comparative}, \citeA{yang2007bayesian}, \citeA{raje2012bayesian}). Recently \citeA{lu2021streamflow} has applied Bayesian LSTM in uncertainty quantification in streamflow prediction. Bayesian Neural Networks were made by approximating the intractable Bayesian Inference. BayesByBackprop \cite{blundell2015weight} is a backpropagation-compatible algorithm to learn probability distribution of a neural network. The same concept is extended to create Bayesian LSTM.

Bayesian Neural Networks can provide us uncertainty on weights by sampling them from a distribution parameterized by trainable variables \cite{esposito2020blitzbdl}. Thus the parameters are sampled using the following equations ($\rho$ and $\mu$ parameterizes the standard deviation and the mean of the samples, respectively.):

For the weights:
\begin{subequations}

\begin{equation}\label{eq:18}
W^{(i)}_{(n)}=N(0,1)*log(1+\rho^{(i)})+\mu^{(i)}
\end{equation}
where $W^(i)$ corresponds to the weight for the $i^{th}$ layer.

\begin{equation}\label{eq:19}
b^{(i)}_{(n)}=N(0,1)*log(1+\rho^{(i)})+\mu^{(i)}
\end{equation}
where $b^{(i)}$ corresponds to the bias for the $i^{th}$ layer.

\begin{algorithm}[]
\SetAlgoLined
Let $\epsilon = N(0,1)$ \\
Let $\theta = (\rho,\mu)$ \\
Let $w = \mu + log(1+e^{\rho})*\epsilon$ \\
Let $f(w,\theta)$ be differentiable relative to its variables.\\
$\Delta_{\mu}= \frac{\delta f(w,\theta)}{\delta w} + \frac{\delta f(w,\theta)}{\delta \mu}$ \\
$\Delta_{\mu}= \frac{\delta f(w,\theta)}{\delta w}    \frac{\epsilon}{1+e^{\rho}}+\frac{\delta f(w,\theta)}{\delta \rho}$ \\
\caption{Training Procedure}
\end{algorithm}

As proposed by \citeA{blundell2015weight}, Kullback-Leibler Divergence \cite{Joyce2011} between priori distribution, $P(w)$ and posteriori empirical distribution, $Q(w|\theta)$ is used as a loss function $f(w,\theta)$ for sampled weights, given their parameters. Following approximation of Kullback-Leibler Divergence used to train the model to speedup the procedure.

\begin{equation}\label{eq:20}
f(w,\theta) \approx \Sigma_{i=1}^{n} log Q(w^{(i)}|\theta) - log P(w^{(i)}) - log P(D|w^{(i)})
\end{equation}
\end{subequations}

where, $w^{(i)}$ denotes the $i^{th}$ Monte carlo sample drawn from $Q(w^{(i)}|\theta)$ and $D$ shows the training data.

\noindent\textbf{Text S4: Gaussian process regression (GPR)}

 It is a non-parametric kernel-based probabilistic model \cite{williams2006gaussian} in which the Gaussian processes assumes model outputs should have a Gaussian joint probability distribution. Researchers have used the GPR for daily \cite{rasouli2012daily} and monthly \cite{sun2014monthly} streamflow forecasting. The details of GPR are described as follows \cite{niu2021evaluating}:

The noise signal in nonlinear regression can be expressed as
\begin{subequations}

\begin{equation}\label{eq:21}
    {y = f(t) + \varepsilon}
\end{equation}
where, $\varepsilon$ $\sim$ $N$(0, $\sigma^2$) is the error. $y$ and $f(t)$ represent response variable and covariate variables, respectively. The covariance function between two samples is given by:
\begin{equation}\label{eq:22}
    k(t,t';\theta) = Cov(f(t),f(t'))
\end{equation}
where, {$\theta$} is the hyper-parameters of the mean function. 
For $n$ data samples, 
\begin{equation}\label{eq:23}
    y_i = f(t_i) + {\varepsilon_i}
\end{equation}
where, $\varepsilon_i$ follows a normal distribution with zero mean and variance, $\sigma^2$. Then, the joint distribution of ${y_i}$ for $i = 1,2,..,n$ will be multivariate normal, and it can be expressed as:
\begin{equation}\label{eq:24}
    {y = (y_1,y_2,...,y_n)^T \sim N_n(u, \psi)}
\end{equation}
where, the elements in the mean matrix can be written as $u_i = u(t_i)$ and the elements of the square matrix $\psi$ are determined by:
\begin{equation}\label{eq:25}
    {\psi_{ij} = Cov(y_i,y_j) = k(t,t';\theta) + {\sigma}^2 \zeta_{ij}}
\end{equation} 
where, $\zeta_{ij}$ is the Kronecker delta. In the new sample, $y^*$ and $t^*$ are the response value, and input variables, respectively, are assumed without losing generality. The conditional distribution of the value of new sample, $y^*$ can be normal to the form shown below:
\begin{equation}
\label{eq:26}
\begin{split}
  E(y^*\|D) = u(t^*) + {\psi^T}(t^*){\psi^{-1}}(y - u) \\
   Var(y^*\|D) = k(t^*,t^*;\theta) + {\sigma^2} - {\psi^T}(t^*){\psi^{-1}}{\psi}(t^*)  \\
  {\psi}(t^*) = (k(t^*,t_1;\theta), k(t^*,t_2;\theta),...,k(t^*,t_n;\theta))  \\
 \end{split}  
\end{equation}
where, $t^*$ is the input variable, $\psi$ is the covariance matrix of $(y_1,y_2,...,y_n)^T$, and $\psi(t^*)$ is the covariance matrix between $f(t^*)$ and $(f(t_1), f(t_2),...,f(t_n))^T$. The noise variance $\sigma^2$, the hyper-parameters $\theta$ from covariance function and $\beta$ parameter from mean function are estimated by solving following equation:
\begin{equation}\label{eq:27}
  l(\theta, \sigma^2, \beta\|D) = \frac{-1}{2}[{log(det(\psi)) + (y - u)^T {\psi^{-1}} (y - u) + nlog(2\pi)]}
\end{equation}
\end{subequations}

\noindent\textbf{Text S5: Support vector regression (SVR)}

The support vector machine (SVM) is developed for classification, and it is extended to regression by \citeA{vapnik1995nature}. The SVR is one the most popular ML approaches for daily (\citeA{dibike2001model},  \citeA{malik2020support}) and monthly \cite{maity2010potential} streamflow predictions. SVRs can efficiently learn the non-linear relationships between predictors(input variables) and predictands (output variables) using the kernel trick, which maps the inputs into high-dimensional feature spaces. The details of SVR are described here as follows \cite{luo2019hybrid}:

The training data of $N$ samples can be shown as $(X_i,d_i)_i^N$ with $X_i$ and $d_i$ are input vector and desired output, respectively. The SVR can be written as:
\begin{subequations}
\begin{equation}\label{eq:28}
    y = W\psi(X) + b 
\end{equation} 
where, $\psi(X)$ denotes a non-linear mapping while $W$ and $b$ are hyperplane and offset, respectively. For finding suitable SVR function $y$, the regression problem can be written as:
For without a penalty  allocation ,
\begin{equation}\label{eq:29}
|d_i - y_i| \leq \varepsilon
\end{equation}
For a penalty allocation,
\begin{equation}\label{eq:30}
|d_i - y_i| > \varepsilon
\end{equation}
The estimated value within the $\varepsilon- $ insensitive tube makes the loss value as zero. The following equation obtains the parameters of regression with C as regularized constant influencing model complexity and the empirical error.
\begin{equation}\label{eq:31}
min\Bigg[\frac{1}{2} ||W||^2 + C\sum_{i=1}^{N}L^{\varepsilon}(y_i,d_i)\Bigg]
\end{equation}
\begin{equation}\label{eq:32}
L^{\varepsilon}(y_i,d_i) = max(0, |y_i - d_i| - \varepsilon)
\end{equation}
The introduction of slack variables $\xi^+$ and $\xi^-$ in equation (31) results in the following optimization problem.
\begin{equation}\label{eq:33}
min\Bigg[\frac{1}{2} ||W||^2 + C\sum_{i=1}^{N}(\xi^+ + \xi^-)\Bigg]
\end{equation}
Subject to: 
${(W\psi(X_i) + b) - d_i \leq \varepsilon + \xi^+_i}$ \\
${d_i - (W^T\psi(X_i) + b) \leq \varepsilon + \xi^-_i}$ \\
$\xi^+ > 0, \xi^- > 0$ \\
The dual form of above optimization problem can be written as:
\begin{equation}\label{eq:34}
max\Bigg[\sum_{i=1}^{N}(\alpha^+_i - \alpha^-_i)y_i - \varepsilon \sum_{i=1}^{N}(\alpha^+_i + \alpha^-_i) - \frac{1}{2}\sum_{i=1, j=1}^{N}(\alpha^+_i - \alpha^-_i)(\alpha^+_j - \alpha^-_j) K(X_i,X_j) \Bigg]
\end{equation}
Subject to: 
$0 \leq \alpha^+_i \leq C$ \\
$0 \leq \alpha^-_i \leq C$ \\
$\sum_{i=1}^{N}(\alpha^+_i - \alpha^-_i) = 0$ \\
\begin{equation}\label{eq:35}
K(X_i,X_j) = \psi(X_i)^T.\psi(X_j)
\end{equation}\\
where, $K(X_i, X_j)$ shows a non-linear kernel function used for mapping inputs at lower dimension to higher dimension linear space. 
\end{subequations}

\noindent\textbf{Text S6: LASSO and Ridge regression}

The Least absolute shrinkage and selection operator (LASSO) regression is proposed by \citeA{tibshirani1996regression}. It has been used for various applications such as identifying key independent variables \cite{alnahit2020quantifying}, integrating with other techniques to screen the influential variables \cite{chu2018monthly}, and selecting variables having a significant impact on given process \cite{chu2020streamflow}. It  minimizes the regression coefficients of the inputs which are not affecting the target variable. The sum of its absolute values constrains the coefficients. LASSO regression effectively avoids the overfitting by penalizing large regression coefficients, which helps in effective feature selection. The coefficient estimates $(\beta)$ in LASSO regression is calculated with ordinary least squares method by defining loss function, $L(\beta)$ as follows \cite{chu2018monthly}:  

\begin{subequations}

\begin{equation}\label{eq:36}
    {L(\beta) = \|Y - X\beta\|}
\end{equation}
The predictors are selected by adding penalty, $\lambda\beta$ with positive regularization parameter, $\lambda$ controlling the bias, and model parsimony:
\begin{equation}\label{eq:37}
    {L(\beta) = \|Y - X\beta\| + \lambda\beta}
\end{equation}
For the $N$ pairs of samples, the coefficient estimates of the LASSO model can be formulated as:
\begin{equation}\label{eq:38}
    \beta = \arg\min \bigg\{\sum_{i=1}^{N}\Big(y_i - \beta_0 - \sum_{j=1}^{k}\beta_j x_{ij}\Big)^2\bigg\}
\end{equation}
Subject to:
\begin{equation}\label{eq:39}
    \sum_{j=1}^{k}\beta_j \leq \lambda
\end{equation}
\end{subequations}

 Ridge regression is applied for monthly streamflow forecasting \cite{lima2010climate}, and its output is used to compare with state-of-art machine learning models (\citeA{xiang2020rainfall},  \citeA{chokmani2008comparison}). The ordinary least square (OLS) method minimizes the sum of the squares of the residuals. Multicollinearity is often encountered when explanatory or input variables are highly correlated, which results in an unconstrained solution. The multicollinearity decreases the performance of OLS. To handle this issue the ridge regression is used in this study. The ridge regression is proposed by  \citeA{hoerl1970ridge}. It modifies the OLS method. It tries to reduce the coefficients of inputs that do not influence the target variable, but it will not make the coefficient zero. The ridge regression does not remove any of the variables. The regression parameters are calculated by \cite{lima2010climate} following equation:
 \begin{subequations}
 \begin{equation}\label{eq:40}
    \hat{\beta} = \arg\min_\beta \bigg\{\sum_{i=1}^{N}(v_i(t) - f_i(x,t))^2 + \delta \sum_{j=1}^{p}\beta_{jt}^2\bigg\}
\end{equation}
where, $f_i(x,t)$, $v_i(t)$ are linear functions and a normally distributed random variable, respectively. The $\delta$ is the ridge parameter that decides the shrinkage of the parameters $\beta_{jt}$. Higher the value of $\delta$ more is the shrinkage.
The solution of ridge regression in matrix form is written as:
 \begin{equation}\label{eq:41}
    \hat{\beta} = \Big(X^TX + \delta I \Big)^{-1} X^Tv
\end{equation}
\end{subequations}
where, $X$ is the input matrix, and $I$ is the identity matrix with dimensions of $p\times p$.

\bibliography{references.bib}
%
%
%


%
%
%
%
%

%
%
\end{article}
\clearpage